\documentclass[10pt,twocolumn,letterpaper]{article}

\usepackage{3dv}
\usepackage{times}
\usepackage{epsfig}
\usepackage{graphicx}
\usepackage{amsmath}
\usepackage{amssymb}

\usepackage{comment}
\usepackage{amsmath}
\usepackage{changepage}
\usepackage{hhline}

\usepackage{booktabs}
\usepackage{multirow}
\usepackage[table,xcdraw]{xcolor}

\usepackage{color}
\usepackage[table]{xcolor}

\definecolor{ben}{rgb}{0.9,0.,0.5}

\definecolor{todo}{rgb}{1.0, 0., 0.}

\usepackage[pagebackref=true,breaklinks=true,letterpaper=true,colorlinks,bookmarks=false]{hyperref}

\threedvfinalcopy 

\def\threedvPaperID{155} 

\ifthreedvfinal\fi
\begin{document}

\title{Wild ToFu: Improving Range and Quality of Indirect Time-of-Flight Depth\\with RGB Fusion in Challenging Environments}

\author{HyunJun Jung$^{1}$, Nikolas Brasch$^{1}$, Ale\v{s} Leonardis$^{2}$, Nassir Navab$^{1}$, Benjamin Busam$^{1}$\\
$^{1}$ Technical University of Munich, $^{2}$ Huawei Noah's Ark Lab\\
{\tt\small \{hyunjun.jung, nikolas.brasch, nassir.navab, b.busam\}@tum.de}
}

\maketitle
\thispagestyle{empty}

\begin{abstract}
Indirect Time-of-Flight (I-ToF) imaging is a widespread way of depth estimation for mobile devices due to its small size and affordable price.
Previous works have mainly focused on quality improvement for I-ToF imaging especially curing the effect of Multi Path Interference (MPI).
These investigations are typically done in specifically constrained scenarios at close distance, indoors and under little ambient light.
Surprisingly little work has investigated I-ToF quality improvement in real-life scenarios where strong ambient light and far distances pose difficulties due to an extreme amount of induced shot noise and signal sparsity, caused by the attenuation with limited sensor power and light scattering.
In this work, we propose a new learning based end-to-end depth prediction network which takes noisy raw I-ToF signals as well as an RGB image and fuses their latent representation based on a multi step approach involving both implicit and explicit alignment to predict a high quality long range depth map aligned to the RGB viewpoint. We test our approach on challenging real-world scenes and show more than 40\% RMSE improvement on the final depth map compared to the baseline approach~\cite{su}. 

\end{abstract}

\section{Introduction}

\begin{figure}[htpb]
    \centering
    \includegraphics[width=\linewidth]{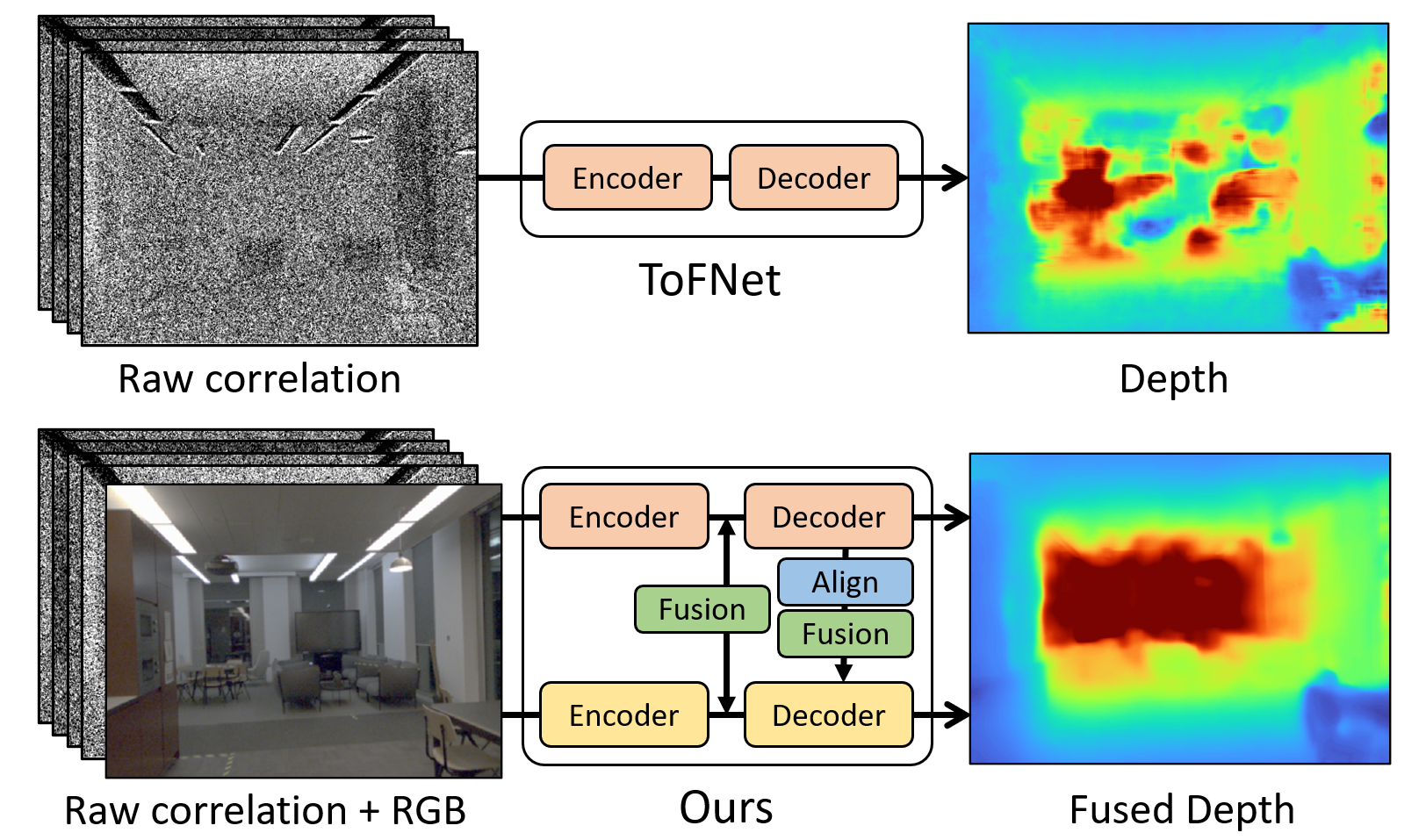}
    \caption{Unlike the baseline approach ToFNet \cite{su} which only uses the raw correlation signals, we leverage an additional RGB input by fusing its latent representation in a multi-step alignment process and achieve a significant improvement of over 40\% RMSE in challenging indoor scenarios.} 
    \label{fig:teaser}
\end{figure}

Depth sensing is a crucial element in modern 3D vision applications such as realistic geometry-aware augmented reality~\cite{holynski2018fast,busam2019sterefo} and an enabling technology for numerous robotics applications~\cite{hwang2020applying,busam2015stereo}.
A variety of cameras and technologies to estimate distances exist.
Prominent passive depth prediction methods such as multi-view stereo suffer in low-textured areas and with limited ambient light while active methods such as structured light struggle in bright ambient light environments and may require multiple consecutive image acquisitions which hamper their real-time use~\cite{geng2011structured}.

Time-of-Flight (ToF) imaging is an active sensing method that can produce real-time depth maps.
For indirect ToF (I-ToF), the scene is illuminated with an amplitude-modulated light typically in the infrared spectrum.
Due to the constant speed of light $c$, depth can be computed from the round-trip time of emitted photons.
To measure the time delay, the sensor uses specific hardware to correlate a number of phase-shifted signals from the emitter diode with the received light reflected from scene geometry. Measuring 4 phase shifts per pixel generates the correlation images $C_i, i \in \{ 1,2,3,4\}$ that are directly connected to the depth map $D$ (see Sec.\ref{sec:tof}).
Complementary to stereo RGB, depth sensing from ToF works well in low-light scenarios and even for textureless objects due to its active illumination.
Since the emitted light intensity decreases quadratically with the distance, ToF depth maps are typically less accurate at further distances.
Ambient light sources can introduce high levels of noise on the correlation maps resulting in missing depth values or incorrect estimates.
Strongly diffuse, absorptive (dark) and reflective materials as well as unwanted multiple reflections can additionally hamper the depth quality.
The complementary advantages and the recent availability even on mobile phones~\cite{noraky2019low} make Time-of-Flight an interesting modality that has recently been studied mainly under ideal environmental conditions, in lab scenarios with low ambient light and on synthetically created images~\cite{FLAT,su,tof_adv,tof_pyramid}.

\paragraph{Overview and Contribution.}

Previous correlation-based ToF pipelines target real applications in close range with controlled external illumination and utilize synthesized pixel-perfect ground truth signals for supervision (see Fig.~\ref{fig:quality_compare}). In this work, we investigate a way to improve the quality and range of single frequency indirect Time-of-Flight for challenging real indoor scenarios such as scenes with distances beyond the manufacturer's maximum range with changing and strong ambient light by leveraging extra RGB input.

Using complementary multi-modal images can be problematic when they are not aligned. Both images can be aligned by calibration only if a high quality depth map is available. However, the quality of an I-ToF depth map is often not accurate enough for this task due to sparsity and depth errors caused by Multi Path Interference (MPI) or Motion Artifacts (MA). Matching algorithms also fail due to a large domain difference between I-ToF and RGB~\cite{tof_book}. Qiu et al.~\cite{tof_align} propose to use a flow based alignment between I-ToF depth and RGB which can be used to align the images to fix low frequency errors such as MPI. However, this approach is constrained to exploit the RGB information as the obtained depth map from the camera already lost information on the problematic low confidence regions during on-camera's pre-processing, leaving no rooms to improve the depth map in those regions jointly with RGB from raw I-ToF signals. While aligning raw signals to the RGB frame using a flow algorithm can be problematic in the presence of extreme sparsity and shot noise typically present in real-life scenarios.

To exploit the richness of information in the raw signal jointly with an unaligned RGB image, we propose a multi step approach which gradually aligns both information while aggregating them at the same time. First, we align both camera inputs with the same intrinsic $K$, then feed them into independent encoders with a low resolution bottleneck (1/32 input resolution). Next, we fuse both feature maps from the bottleneck assuming both of them are weakly aligned to predict an improved depth at low resolution. Then we repeat the process estimating up-scaled depth maps until recovering the full resolution by 1) aligning I-ToF feature maps to RGB feature maps by warping with an intermediate depth and pose 2) applying a fusion module 3) hierarchical upsampling and depth estimation, which at the end predicts high quality depth on the RGB viewpoint. In the evaluation we show that our proposed method resolves the alignment issue while significantly improving over baseline approaches.


To this end, our main contributions are:
\begin{enumerate}
  \setlength{\itemsep}{0pt} 
  \setlength{\parskip}{0pt}
    \item We tackle I-ToF in a realistic and challenging scenario which involves strong ambient light and at ranges even beyond the manufacturer's maximum depth.
    \item We leverage unaligned RGB information jointly with I-ToF at low resolution while resolving their alignment issue with a special architecture capable of producing high resolution depth outputs.
    \item We provide insight for fusion strategy between spatially separated sensors by extensive ablation studies.
\end{enumerate}

\section{Related Work}
\label{sec:related}

Previous works related to I-ToF depth correction can be separated into two branches. The first is directly connected to the I-ToF modality which mostly targets to improve specific noise patterns in the modality such as MPI from either raw signal or estimated depth from the camera. The other branch is not necessarily limited to I-ToF, but focuses on completing depth with RGB input in case of sparsity which can be applied to I-ToF depth as it often suffers from sparsity due to its limited power and scattering.

\subsection{Indirect Time-Of-Flight}

\paragraph{From raw correlations}
Early correction approaches using raw correlation images investigate mainly MA and MPI which are commonly addressed sources of I-ToF depth error. A few works~\cite{schmidt2011high,samsung_invariant} focus on constraints in the correlation images to improve MA, and other works \cite{fuchs2010multipath,fuchs2013compensation,jimenez2014modeling,dorrington2011separating,godbaz2012closed,kirmani2013spumic} specifically target MPI artefacts. Some works focus on improving the overall range of the depth map by resolving the periodicity in the calculated phase (phase unwrapping, Sec.\ref{sec:tof}) using intensity \cite{intensity_unwrap}, multi-frequency modulation \cite{multifreq_unwrap} or signal distribution \cite{distribution_unwrap}.

While the aforementioned ToF approaches offer the possibility to easily exchange modules for denoising, phase unwrapping and MPI / MA correction, they work only independently and suffer from error propagation. Some scholars \cite{su,FLAT} address this problem by applying learning based methods on the raw correlation data.
Su et al.~\cite{su} understood the benefit of training with direct sensor input and use dual-frequency raw correlation images for an encoder-decoder network to predict depth in an end-to-end approach, while Guo et al.~\cite{FLAT} use a window-based approach with a Kernel Prediction Network (KPN)~\cite{KPN} to pre-process the raw correlation images before extracting the depth. They both synthetically create a training dataset with a physics based renderer (Fig.\ref{fig:corr_to_depth}), which is capable of generating physical noise such as MPI and scattering as well as Gaussian and Poisson noise. However, the rendering resembles controlled lab scenarios which have limited ambient light and close distance objects which does not reflect challenging real world scenarios as exemplified in Fig.\ref{fig:Sparsity}. Although our work shares the similarity with these two works by taking raw correlation images as input and also uses a learning based approach, we rather focus on a more realistic and challenging setup, such as single frequency raw correlation images in uncontrolled and long distance indoor scenes to which previous works cannot generalize well.




\paragraph{From I-ToF depth}
Another series of works focuses on directly improving the depth map obtained by an I-ToF camera.
At this stage the signal may have already undergone linear and nonlinear models from either hardware or software of the camera and information in areas with low confidence measures may be lost. 
Some scholars investigate I-ToF depth improvement as a post processing stage~\cite{son2016learning, deeptof, tof_pyramid} focusing on removing MPI, MA, and shot noise while others fuse ToF with complementary information from an RGB camera~\cite{tof_align} or use generative adversarial networks to bridge the gap between synthetic and real data~\cite{tof_adv}. These works are evaluated in relatively optimized scenarios where the main error source is MPI or MA. However, as shown in Fig.\ref{fig:quality_compare} and Fig.\ref{fig:Sparsity}, I-ToF depth in real scenarios with strong ambient light suffers mainly from sparsity and an extreme amount of shot noise.

\subsection{Depth Completion}
The depth map obtained from the depth camera in general contains invalid pixels due to the hardware limitations. Although not ToF specific, there has been recent progress regarding the completion of invalid pixels especially for LiDAR depth jointly with RGB input, which shares similar principles with our work. Common approaches treat the sparse depth as a single channel image input for a network \cite{van2019sparse,guidenet,fcfc_lidar}, while more geometric approaches being developed recently treat sparse depth in different form other than a single channel image. Hu et.al.~\cite{pe_lidar} use an extra analytical feature map calculated by back-projecting coordinates with camera intrinsics and sparse depth in multiple layers and Chen et.al.~\cite{acm_lidar} treat the sparse depth map as a point cloud to apply point-based networks. Applying those works directly on raw I-ToF correlation with RGB would be problematic. Geometric approaches are not feasible by design as the raw signal requires more steps to be a used as distance measures as well as its strong noise. And different to LiDAR, the present misalignment between the raw signal and RGB can cause additional artefacts around object boundaries and depth discontinuities. In Sec.\ref{sec:results} we show the impact of the misalignment on the final depth prediction.

\section{Time-of-Flight Sensing}
\label{sec:tof}
In this section we will briefly introduce the background of Time-of-Flight sensing and its possible sources of error.

\subsection{Depth Acquisition from I-ToF}\label{sec:background}

\paragraph*{Correlation Images.}

\begin{figure}[tpb]
    \centering
    \includegraphics[width=\linewidth]{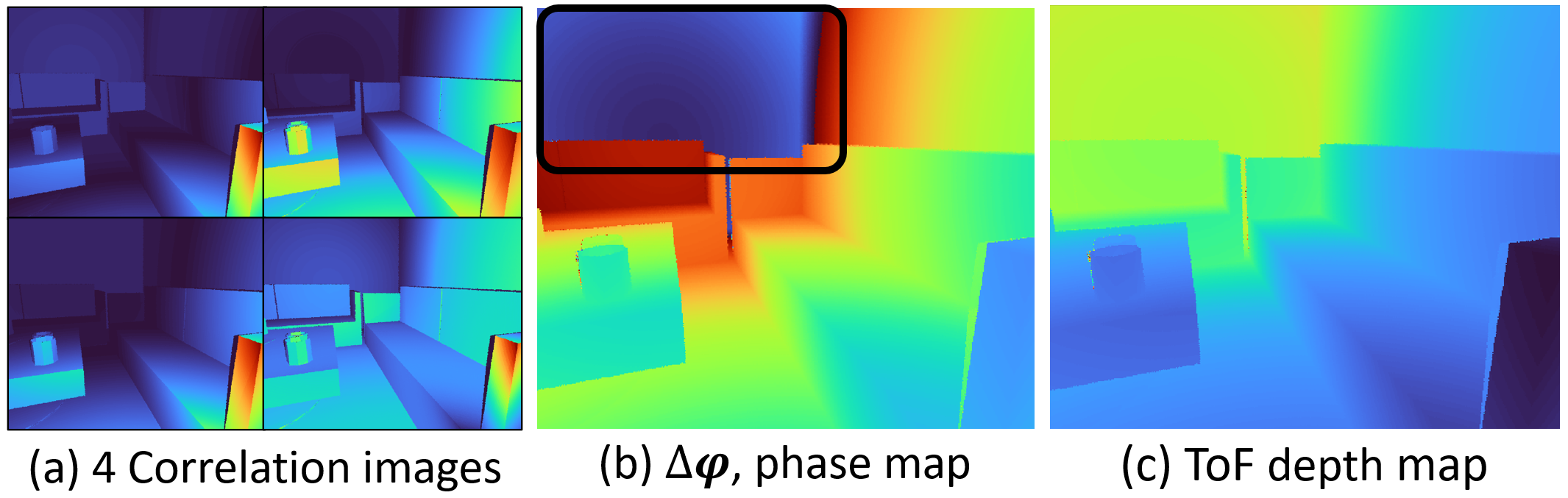}
    \caption{Synthetic I-ToF images from \cite{FLAT}. Each of the 4 correlation images (a) is function of the phase map. Note that the phase values drop suddenly in the marked area of the phase map (b) due to the phase wrapping. The full range depth map (c) can be obtained after phase unwrapping.}
    \label{fig:corr_to_depth}
\end{figure}

When an Indirect Time-of-Flight (I-ToF) camera measures depth, it first emits an amplitude modulated infrared (IR) signal and then measures the returned signal which contains phase shift information. To recover the phase shift, cross-correlation is used. Cross correlation between two signals \(f(t)\) and \(g(t)\) is defined as:
\begin{equation}
(f\otimes g)(\varphi)=\lim_{T\rightarrow \infty}\frac{1}{T}\int_{\frac{-T}{2}}^{\frac{T}{2}}f(t)\cdot g(t+\varphi)dt
\label{eq:crosscorr}
\end{equation}
When we plug our returned signal \(r(t)\) with amplitude \(a\) and phase shift \(\varphi\) into \(g\) and demodulation signal \(d(t)\) with control phase \(\tau\) into \(f\) of Eq~\ref{eq:crosscorr},
\begin{equation}\label{eq:returned}
r(t) = 1 + a\cdot \cos(\omega t - \varphi),\quad
d(t) = \cos(\omega t - \tau)
\end{equation}
we can obtain the function of both phase shift \(\varphi\) and control phase \(\tau\) with
\begin{equation*}
(r\otimes d)(\varphi)=\lim_{T\rightarrow \infty}\frac{1}{T}\int_{\frac{-T}{2}}^{\frac{T}{2}}[\cos(\omega t - \tau)]\cdot[1 + a\cdot \cos(\omega t - \varphi)]dt
\end{equation*}
\begin{equation}
= \frac{a}{2}\cdot \cos(\varphi +\tau)
\end{equation}
With known control phases \(\tau\) $\in$ [\(0^{\circ},90^{\circ},180^{\circ}\,270^{\circ}\)], we retrieve functions of \(\varphi\) as
\label{eq:correlation}
\begin{equation}
\begin{aligned}
&C_{1}=c_{0}=\frac{a}{2}\cos(\varphi),\  &C_{2}=c_{180}=-\frac{a}{2}\cos(\varphi),\\ &C_{3}=c_{90}=\frac{a}{2}\sin(\varphi),\  &C_{4}=c_{270}=-\frac{a}{2}\sin(\varphi).
\end{aligned}
\end{equation}
These constitute the four raw correlation values which are measured by an I-ToF camera for each pixels in the image and result in four correlation maps as shown in Fig.\ref{fig:corr_to_depth}(a).

\paragraph*{Phase calculation and unwrapping.}
As correlation images are sine functions of phase shift \(\varphi\), the arc-tangent formula can be used to extract \(\varphi\). However, it can be only measured up to 0-2\(\pi\) range due to the nature of the periodic signal. The signal \(\varphi\) with its 2\(\pi\)-ambiguity is called the wrapped phase (see Fig.\ref{fig:corr_to_depth}(b)).

\begin{equation}
\varphi_{wrapped} = \arctan \left( \frac{C_{3}-C_{4}}{C_{1}-C_{2}} \right)
\label{eq:arctan}
\end{equation}
To obtain the full range of the depth map, it is required to find the correct multiplier for 2\(\pi\). This disambiguation is called phase unwrapping.

\label{eq:unwrap}
\begin{equation}
\varphi = \varphi_{wrapped} + 2n\pi \quad (n\in \{0,1,2...\})
\end{equation}
The phase unwrapping factor can be calculated either explicitly by non-learning based approaches such as by using amplitude \cite{intensity_unwrap}, multi frequency modulation signal \cite{multifreq_unwrap}, signal distribution \cite{distribution_unwrap}, or implicitly by learning based approaches such as \cite{su}. However, due to the high amount of noise under real lighting conditions the intensity and distribution methods do not work well. In Sec.\ref{sec:results} we show that in the case of a single frequency I-ToF signal with high noise, the main source of depth error comes from the confusion of the unwrapping factor.

\paragraph*{Depth conversion.}
After the phase \(\varphi\) is unwrapped, the distance \(D\) can be obtained with a conversion formula using the speed of light \(c\) and the modulation frequency \(\omega\) as illustrated in  Fig.\ref{fig:corr_to_depth} (c) with
\begin{align}\label{eq:atan}
D = \frac{c}{2\omega}\cdot\frac{\varphi}{2\pi}
\end{align}

\subsection{Time-of-Flight's Source of Error}
The acquisition process of I-ToF cameras introduces three main sources of error which we briefly describe.

\paragraph*{Multi path interference and motion artefact.}
The superposition of return signals with different \(\varphi\) in one pixel is one of the main sources of I-ToF error. It happens when the returned signal contains multiple reflection paths (MPI) or sudden changes of the returned signal \(\varphi\) due to the camera motion (MA). In both cases, the final correlation signals become superimposed with signals of different \(\varphi\) and Eq.\ref{eq:arctan} in turn fails to provide the accurate phase. Aforementioned artefacts can be characterised as low frequency noise around the corner (MPI) and over/under-shooting around the object boundary (MA). However, we will identify that these artefacts are not the main problem in challenging real scenes.

\begin{figure}[tpb]
    \centering
    \includegraphics[width=\linewidth]{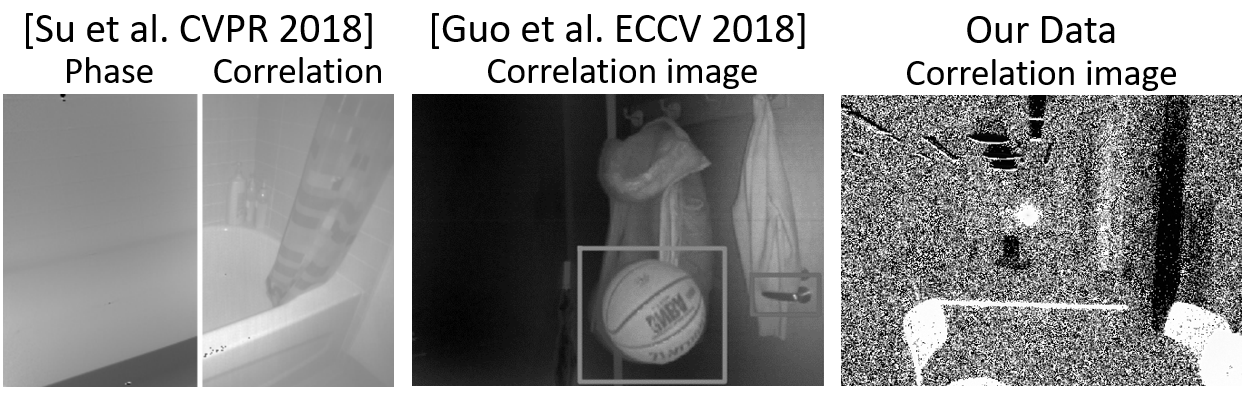}
    \caption{Comparison of input correlation noise on real data from Su et al.~\cite{su}, Guo et al.~\cite{FLAT}, and us. The significantly higher noise level of our correlation images is immanent.}
    \label{fig:quality_compare}
\end{figure}

\paragraph*{Sparsity and shot noise.}
The raw correlation images from the I-ToF camera rely on the returned IR signal from reflections. Any factors causing scattering of the IR signal can degrade the quality of the raw correlation images, such as the surface normal or the material properties of objects in the scene. Hansard et al.~\cite{tof_book} shows that the amplitude of the return signal decreases and the depth error increases as the surface normal of the object deviates from the optical axis of the camera and when the object material is reflective due to a scattering effect. In real-life scenarios, when the aforementioned regions are combined with strong ambient light, the signals become highly noisy as shown in Fig.\ref{fig:Sparsity}. Non-static ambient illumination can additionally cause incorrect measurements of return signals in practice. To reduce the correlation deterioration by ambient light, previous investigations focused on ideal environmental conditions during data acquisition while we specifically target less constraint setups which manifests in significant differences in the input data as shown in Fig.\ref{fig:quality_compare}.

\section{ToF Fusion}

\begin{figure*}[htbp]
        \centering
        \includegraphics[width=\linewidth]{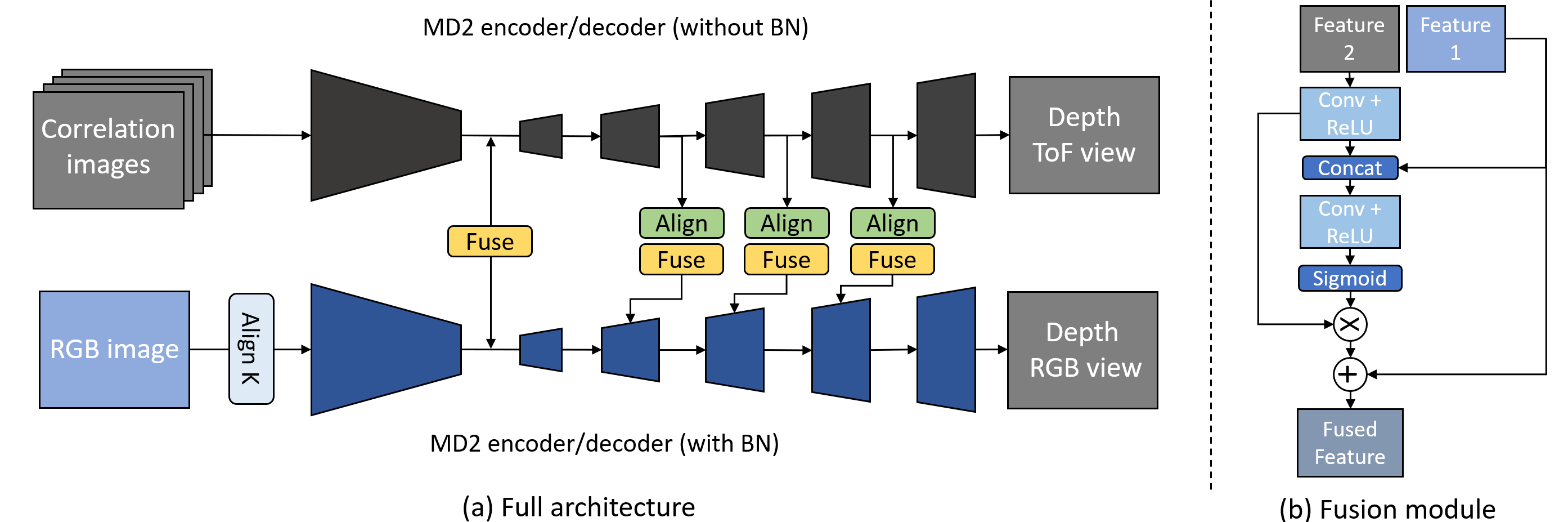}
\caption{Details of our network architecture. While the lower branch in (a) follows~\cite{monodepth2}, the upper branch in uses~\cite{monodepth2} without Batch-Norm and with a 4-channel input. The fusion in the bottleneck combines both feature maps bidirectionally without any alignment, while higher resolution fusions are performed unidirectionally from I-ToF to the RGB branch with intermediate depth and extrinsic calibration between the cameras. All fusion steps use the fusion module described in (b).}
\label{fig:full_architecture}
\end{figure*}

Our pipeline uses an encoder-decoder style fully convolutional network with raw I-ToF input similar to \cite{su}. However, significant changes are made to deal with the extreme amount of noise in the raw signal and to leverage the unaligned RGB image.

\subsection{Network Backbone}
We use a two branch encoder-decoder network architecture with independent weights to predict the depth map from raw I-ToF input signal as well as RGB input. The overall architecture is illustrated in Fig.\ref{fig:full_architecture}. Different from \cite{su} where the 1/4 resolution feature map serves as bottleneck, we use the same encoder-decoder architecture with 1/32 resolution bottleneck as in \cite{monodepth2} for both RGB and I-ToF branch separately, but with additional feature sharing in the bottleneck and decoder layers with our custom fusion module (Sec.\ref{sec:fusion_module}). For the I-ToF branch, we modify the architecture by expanding the input channel dimension while removing all Batch-Norm layers so that the network can process 4 correlation images as input and keep their channel wise relation unchanged as in \cite{su}. There are three reasons behind our choice of the architecture:
\begin{enumerate}
  \setlength{\itemsep}{0pt} 
  \setlength{\parskip}{0pt}
    \item The 1/32 resolution bottleneck helps to extract features in a more global context, which already improves the depth prediction in areas of unreliable I-ToF.
    \item Misalignment between RGB and I-ToF is less problematic at 1/32 resolution, making fusion possible between both feature maps without any alignment.
    \item Intermediate depth from the decoder can be used to align intermediate feature maps from one to the other. 
\end{enumerate}

\subsection{Intrinsic Alignment} \label{subsec:intrinsic_align}
\begin{figure}[htpb]
    \centering
    \includegraphics[width=\linewidth]{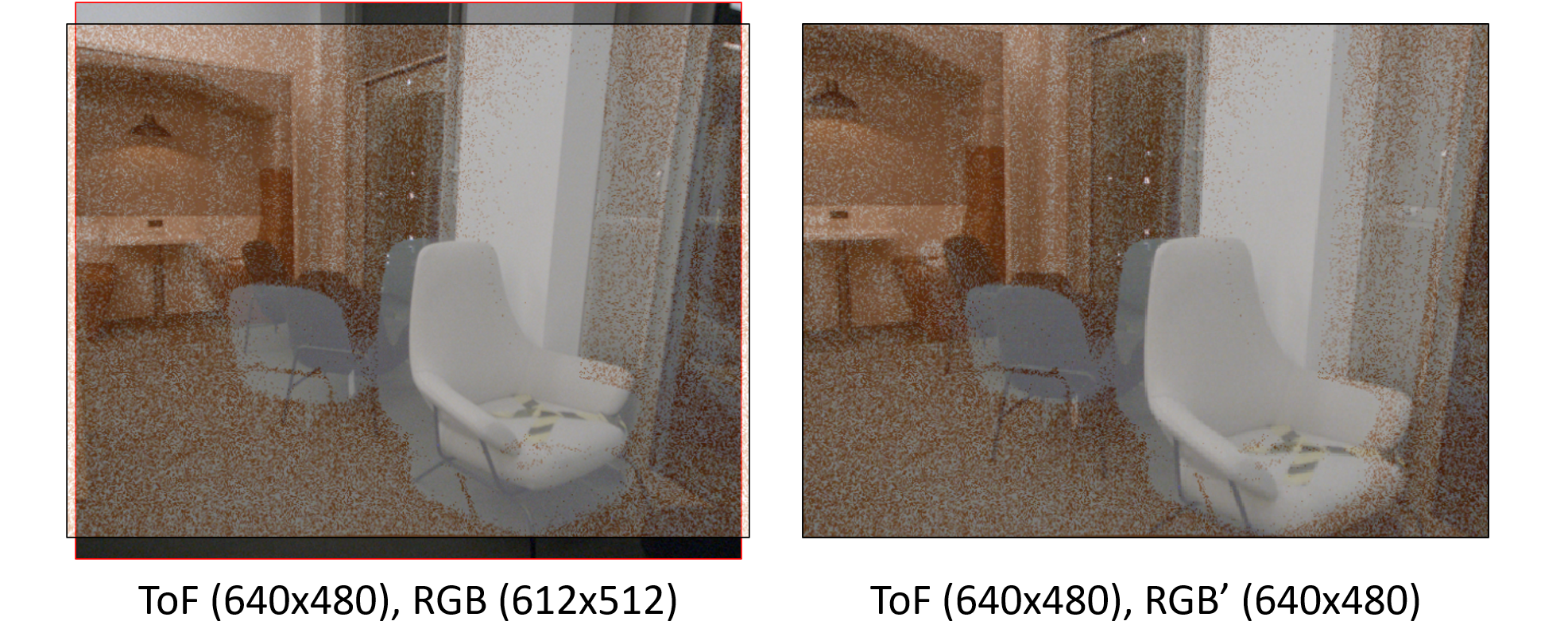}
    \caption{Two input images overlaid before (Left) and after (Right) intrinsic alignment. After the alignment, both images share equal resolution and the scene in the field of view only involves displacement, no scaling.}
    \label{fig:intrinsic_alignment}
\end{figure}

Although the 1/32 resolution bottleneck of ResNet-18~\cite{resnet} helps to ease the misalignment between I-ToF and RGB camera, the difference in camera intrinsic $K$ between the sensors can cause issues due to different resolution and field of view (FOV) during the fusion. To overcome this, we warp from the image $I_1$ with the larger field of view (RGB) onto the image $I_2$ with the smaller FOV (I-ToF) using the camera intrinsics $K_1, K_2$ of both cameras as depicted in Eq.\ref{eq:intrinsic_alignment} such that both branches have the same resolution and least amount of invalid pixel (see Fig.\ref{fig:intrinsic_alignment}).
\begin{equation}
\begin{aligned}
    (x_{2},y_{2},1)^{T} = K_{2}K_{1}^{-1}(x_{1},y_{1},1)^{T} \\
    I_{1 \rightarrow 2}(x_{2},y_{2}) = I_{1}(x_{1},y_{1})
\end{aligned}
\label{eq:intrinsic_alignment}
\end{equation}

\subsection{Fusion Module}
\label{sec:fusion_module}
Aligning images from different modalities is an important step before the fusion. Qiu et.al.~\cite{tof_align} use a flow based network to align from I-ToF to RGB. However, this cannot be used in our setup due to the sparsity and strong noise which makes flow estimation difficult. Explicit alignment before fusion is not possible either with only RGB depth or I-ToF depth as I-ToF depth is unreliable due to the noise and phase unwrapping error, while RGB depth prediction produces poor depth when the scene has little texture or photometrically-changing geometry to infer depth from. To overcome this problem, we utilize a two step approach. First, we use encoders with low resolution bottleneck and then fuse both feature maps to predict improved intermediate depth maps at low resolution. Then we warp the I-ToF feature map onto the RGB reference frame with intermediate depth predictions and camera extrinsics and apply fusion inside the RGB decoder.
\begin{equation}
    (x_{2},y_{2},1)^{T} = K_{2}h((Rd_{1}K_{1}^{-1}(x_{1},y_{1},1)^{T}+t))
\end{equation}
Where camera 1 is I-ToF and 2 is RGB. \(R,t\) denote the extrinsics, \(d_{1}\) intermediate depth prediction, and \(h\) indicates homogenization. We iterate this for the RGB decoder at resolution (1/16\(\rightarrow\)1/8), (1/8\(\rightarrow\)1/4), (1/4\(\rightarrow\)1/2). The detailed structure of the entire network is described in our supplementary material. Passing the features from I-ToF to RGB has the advantage of relying on the finest depth possible in the RGB viewpoint at the end which is then ready for applications such as augmented reality. 

The fusion module structure is shown in Fig.\ref{fig:full_architecture} (b). As mentioned, both features are assumed to be aligned either implicitly (at 1/32 res.) or explicitly (with fixed extrinsics between the two sensors for intermediate depth). To convert each feature map to the other, we use a Conv-layer followed by a ReLU-activation. To iterate on the confidence of the converted feature map, an additional Conv-layer with ReLu is used after concatenation with the additional feature map. A Sigmoid-activation is applied to normalize the output to the range $[0,1]$. This attention weight is applied to the converted feature map before it is added to the other feature map.


\begin{figure*}[t]
        \centering
        \includegraphics[width=\linewidth]{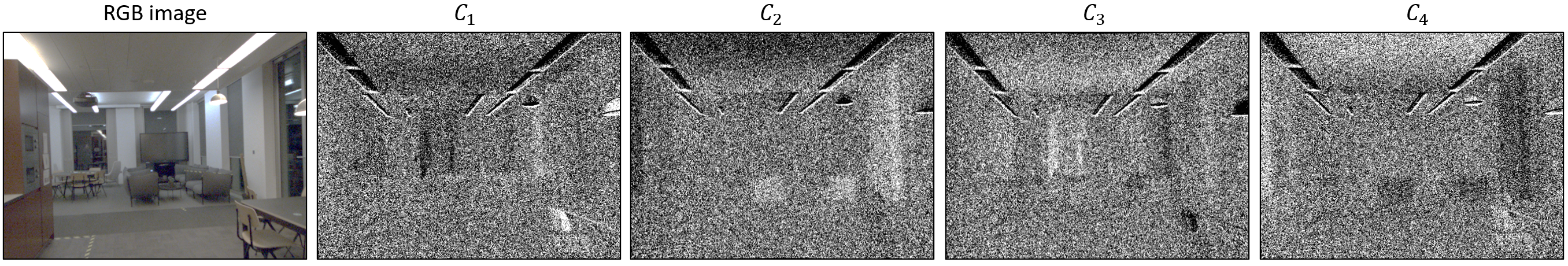}
\caption{Example acquisition of a scene (RGB image) with corresponding raw I-ToF measurements as correlation images $C_i, i \in \{ 1,2,3,4\}$. Due to ambient light and distant regions, the significant amount of shot noise is clearly visible.}
\label{fig:Sparsity}
\end{figure*}

\subsection{Loss Formulation}
We formulate our loss as a fully supervised multi-scale $\mathcal{L}_1$ loss in both the RGB and I-ToF branch with additional edge aware smoothness~\cite{monodepth2} in the RGB branch. For $\mathcal{L}_1$, we adapted the multi-scale approach from \cite{monodepth2}, where each intermediate depth output with resolution \(i\) is upsampled in both branches to the full resolution \(d^i\) which is then compared to the ground truth \(\tilde{d}\). The loss term reads as
\begin{equation}\label{eq:fusion_loss1}
\mathcal{L}_1 = \sum_{i}\left(\sum|d_{tof}^{i} - \tilde{d}_{tof}| + \sum|d_{rgb}^{i} - \tilde{d}_{rgb}|\right)
\end{equation}
Edge aware smoothness is only applied on the RGB image \(I_{rgb}\). Su et al.~\cite{su} apply this loss on the I-ToF branch by treating the I-ToF amplitude as a grey scale image. Due to sparsity and shot noise present in our data, we do not include such a term. The smoothness loss is calculated with
\begin{equation}\label{eq:fusion_loss2}
\mathcal{L}_s = \sum|\partial_{x}d_{rgb}|e^{-|\partial_{x}I_{rgb}|} + |\partial_{y}d_{rgb}|e^{-|\partial_{y}I_{rgb}|}
\end{equation}
Our final loss is a weighted combination of \(\mathcal{L}_1\) and \(\mathcal{L}_s\):
\begin{equation}\label{eq:fusion_loss3}
\mathcal{L} = \mathcal{L}_1 + \lambda_{s}\mathcal{L}_s
\end{equation}

\section{Dataset and Training Details}\label{sec:training}

In this section, we introduce the dataset we used for both training and evaluation as well as the detailed setup for the training
\paragraph*{Training Dataset}
For our work, we leverage the indoor sequences from the multi-modal dataset~\cite{CroMo} which constitutes ToF acquisitions under challenging light and scene conditions as illustrated in Fig.\ref{fig:Sparsity}.
The dataset comprises synchronised image sequences capturing multiple modalities, at video-rate across various real-world indoor and outdoor scenes.
A custom camera rig is constructed in order to capture synchronised data including i-ToF correlations from a Lucid HLS003S-001 camera (LUCID Vision Labs, Canada).
The used RGB images are acquired with a resolution of $612{\times}512$~px while i-ToF captures 4 channel raw correlation images of $640{\times}480$~px resolution with $8$ bit depth.
A structured light sensor (RealSense D435i, Intel, USA) acquires active IR stereo depth maps in the RGB image viewpoint that serve as ground truth.
Intrinsic and extrinsic camera parameters are calibrated via multi-view graph bundle adjustment and provided together with the images.
The indoor sequences are composed of 4 individual video sequences with around 800 frames each, recorded on one floor of large building with different trajectories and viewpoints.
We use 3 full sequences for training which sums up to a total of 2230 frames and leverage the provided test split in the remaining sequence for validation.
\paragraph*{Training Setup}
Before the training we preprocess both ground truth and input images. The ground truth is processed in two different ways to align to both RGB and I-ToF coordinate frames. We change the intrinsics of the ground truth in the same way as mentioned in Sec.\ref{subsec:intrinsic_align} for the RGB branch. For I-ToF supervision, we use forward warping to bring the ground truth depth onto the I-ToF viewpoint using the provided extrinsics. For data augmentation, we apply a sequence of brightness/contrast/saturation/hue transformations for RGB input. We use the pretrained ResNet-18~\cite{resnet} encoder provided by PyTorch \cite{pytorch} for our RGB branch, while the I-ToF branch is trained from scratch. We use the ADAM optimizer for the entire training with an initial learning rate of \(10^{-4}\) which decays \(0.95\) every 1000 iterations. For all setups in Tab.\ref{tab:quantitative}, we train 80 epochs (approx. 12 hours using an Nvidia RTX 3090 GPU) with a loss weighting factor of \(\lambda_{s}=1.0\). We evaluate per epoch and used the result of best performing epoch for our tables.


\begin{figure*}[htpb]
    \centering
    \includegraphics[width=\linewidth]{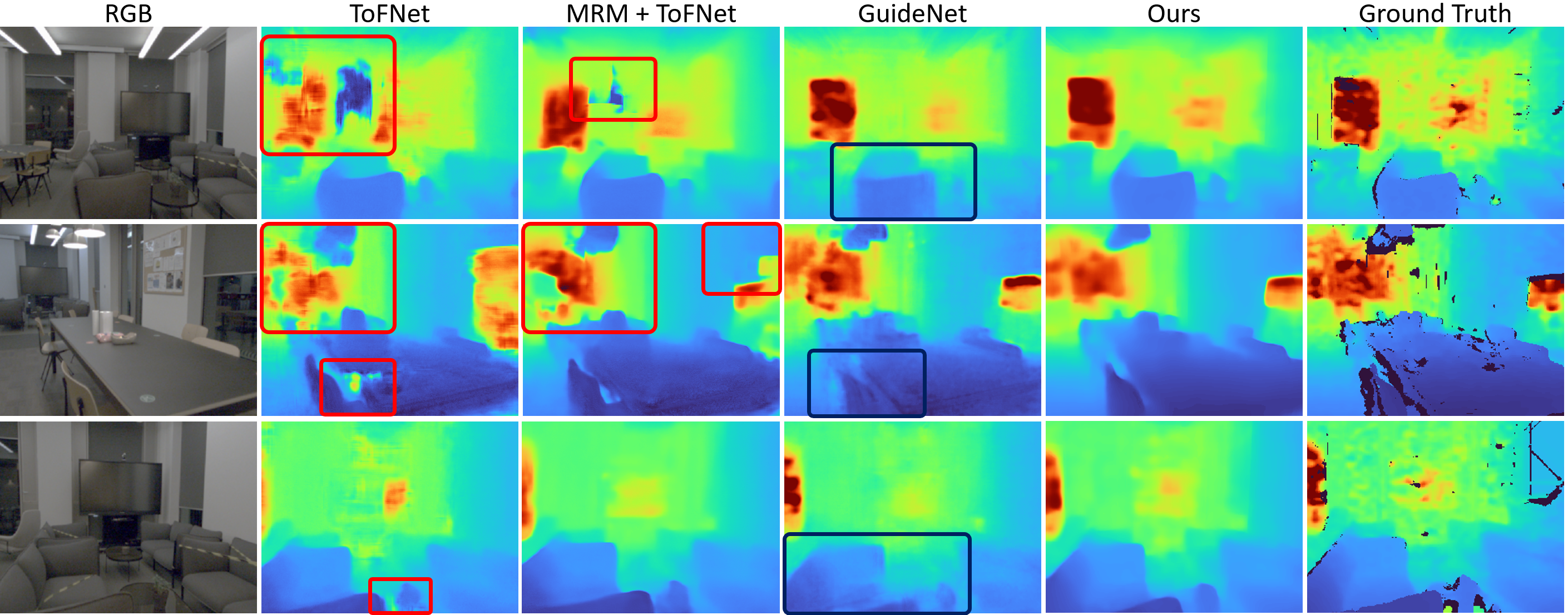}
    \caption{Qualitative comparison of ToFNet~\cite{su}, GuideNet~\cite{guidenet} and our pipeline.  With or without pre-filtering (MRM~\cite{FLAT}), ToFNet suffers from strong artefacts caused by a wrong unwrapping factor (marked red), which results in large regional errors and horizontal/vertical line artefacts. GuideNet \cite{guidenet} resolves this issue with RGB information, however unaligned RGB causes ghosting effects around the discontinuities (marked blue). Our method does not suffer from these artefact and achieves the best result. Note that corresponding correlation images of each RGB image are shown in the supplementary material.}
    \label{fig:qualitative}
\end{figure*}

\begin{table*}[]
\centering
\begin{tabular}{cccccccc}
\hline
\textbf{Backbone}            & \textbf{Input}                      & \textbf{RMSE}                & \textbf{Rel(abs)}             & \textbf{Rel(sqr)}             & \(\mathbf{{1.25}^{1}}\) & \(\mathbf{{1.25}^{2}}\) & \(\mathbf{{1.25}^{3}}\) \\ \hline
ToFNet~\cite{su}                       & Raw I-ToF                           & 0.99                         & 0.13                          & 0.23                          & 87.5                             & 94.6                             & 97.1                             \\
MRM~\cite{FLAT} + ToFNet~\cite{su}                       & Raw I-ToF                          & 0.84                         & 0.098                          & 0.17                         & 91.8                           & 96.5                             & 97.9                             \\
GuideNet~\cite{guidenet} & Raw I-ToF + RGB & 0.61                         & 0.10                          & 0.091                         & 91.2                             & 97.7                             & \cellcolor[HTML]{FFFC9E}99.3     \\
Ours                         & Raw I-ToF + RGB                     & \cellcolor[HTML]{FFFC9E}0.56 & \cellcolor[HTML]{FFFC9E}0.073 & \cellcolor[HTML]{FFFC9E}0.069 & \cellcolor[HTML]{FFFC9E}94.6     & \cellcolor[HTML]{FFFC9E}98.4     & \cellcolor[HTML]{FFFC9E}99.3     \\ \hline
\end{tabular}
\caption{Quantitative comparison between ToFNet~\cite{su}, MRM~\cite{FLAT}+ToFNet~\cite{su}, GuideNet~\cite{guidenet} and our pipeline on the dataset~\cite{CroMo}. (RMSE, Rel: The lower the better, \(1.25^{n}\): The higher the better)}
\label{tab:quantitative}
\centering
\end{table*}

\section{Experiments} \label{sec:results}

For evaluation, we first compare our method to two state of the art methods: ToFNet (raw I-ToF to depth \cite{su}) and GuideNet (RGB guided depth densification \cite{guidenet}), and then justify pipeline components with an ablation study.

\subsection{Quantitative and Qualitative Evaluation}

To quantitatively evaluate our results, we calculate 4 depth metrics suggested in \cite{eigen2014depth} : Root Mean Square Error (RMSE), absolute and square relative error (Rel(abs), Rel(sqr)), and depth accuracy with threshold (\({1.25}^{i}\)).  We first train ToFNet~\cite{su} and GuideNet~\cite{guidenet} from scratch on the multi-modal dataset~\cite{CroMo}. To optimize performance on GuideNet for this data, we make three adjustments to their pipeline: We remove Batch-Norm layers on the depth branch, increase the input dimension to 4, and also use intrinsic alignment (Sec.\ref{subsec:intrinsic_align}) to resolve resolution differences. The training setups follow the original implementation. A quantitative evaluation is shown in Tab.\ref{tab:quantitative} and qualitative results are depicted in Fig.\ref{fig:qualitative}.

In our experiments, ToFNet~\cite{su} achieves the worst overall results among the three methods. Fig.\ref{fig:qualitative} indicates that the largest errors come from the phase unwrapping (second column, marked red). As mentioned in Sec.\ref{sec:tof}, the I-ToF signal has a periodicity and its phase (or depth) is wrapped by 2\(\pi\). This requires extra information to obtain the full range (phase unwrapping). Su et.al applied ToFNet to a relatively clean signal (Fig.\ref{fig:quality_compare}) using dual frequency modulation as input such that the network learns the unwrapping factor easily both analytically and geometrically. This is not the case with the dataset~\cite{CroMo} where I-ToF comes as a single frequency and is sparse and noisy. Improving the quality of the input signal by applying pre-filtering network~\cite{FLAT} removes some of the artefacts, but not completely (the second col in Fig.\ref{fig:qualitative}). GuideNet~\cite{guidenet} achieved a huge improvement over ToFNet especially regarding phase unwrapping, showing the importance of leveraging dense RGB information in this scenario. The mid column of Fig.\ref{fig:qualitative} shows that the unwrapping related artefacts no longer exist in the prediction. However, the unaligned RGB input causes confusion around depth discontinuities which result in blurry predictions and ghosting effects (marked blue). Our proposed method resolves the misalignment issue in multiple steps and achieves the best result both quantitatively and qualitatively. The last column in Fig.\ref{fig:qualitative} shows that our proposed method resolves both phase unwrapping and RGB alignment.

\begin{table*}[ht]
\centering
\begin{tabular}{c|c|c|cccccc}
\hline
\textbf{Backbone}                                                              & \textbf{Fusion}                                                                             & \textbf{Branch} & \textbf{RMSE}                & \textbf{Rel(abs)}             & \textbf{Rel(sqr)}             & \(\mathbf{{1.25}^{1}}\) & \(\mathbf{{1.25}^{2}}\) & \(\mathbf{{1.25}^{3}}\) \\ \hline
                                                                               & No                                                                                          & I-ToF           & 0.99                         & 0.13                          & 0.23                          & 87.5                             & 94.6                             & 97.1                             \\ \cline{2-9} 
                                                                               &                                                                                             & I-ToF           & 0.75                         & 0.098                         & 0.14                          & 92.1                             & 97.0                             & 98.0                             \\
\multirow{-3}{*}{\begin{tabular}[c]{@{}c@{}}ToFNet~\cite{su} \end{tabular}} & \multirow{-2}{*}{\begin{tabular}[c]{@{}c@{}}Bottleneck\\ (implicit align)\end{tabular}}            & RGB             & 0.80                         & 0.11                          & 0.15                          & 88.6                             & 96.0                             & 98.1                             \\ \hline
                                                                               &                                                                                             & I-ToF           & 0.77                         & 0.09                          & 0.14                          & 91.9                             & 96.7                             & 98.4                             \\
                                                                               & \multirow{-2}{*}{No}                                                                        & RGB             & 0.75                         & 0.16                          & 0.17                          & 78.5                             & 95.0                             & 98.6                             \\ \cline{2-9} 
                                                                               &                                                                                             & I-ToF           & 0.58                         & 0.079                         & 0.095                         & 94.0                             & 98.1                             & 99.1                             \\
                                                                               & \multirow{-2}{*}{\begin{tabular}[c]{@{}c@{}}Bottleneck\\ (implicit align)\end{tabular}}           & RGB             & 0.57                         & 0.076                         & 0.071                         & 94.3                             & \cellcolor[HTML]{FFFC9E}98.4     & \cellcolor[HTML]{FFFC9E}99.3     \\ \cline{2-9} 
                                                                               &                                                                                             & I-ToF           & 0.68                         & 0.088                         & 0.12                          & 92.6                             & 97.6                             & 99.0                             \\
                                                                               & \multirow{-2}{*}{\begin{tabular}[c]{@{}c@{}}Bottleneck\\ (attention)\end{tabular}}            & RGB             & 0.68                         & 0.11                          & 0.11                          & 88.2                             & 97.3                             & 99.0                             \\ \cline{2-9} 
                                                                               &                                                                                             & I-ToF           & 0.57                         & 0.076                         & 0.093                         & 94.4                             & 98.2                             & 99.2                             \\
\multirow{-8}{*}{\begin{tabular}[c]{@{}c@{}}MD2~\cite{monodepth2} \end{tabular}} & \multirow{-2}{*}{\begin{tabular}[c]{@{}c@{}}All resolutions\\ (implicit + explicit align)\end{tabular}} & RGB             & \cellcolor[HTML]{FFFC9E}0.56 & \cellcolor[HTML]{FFFC9E}0.073 & \cellcolor[HTML]{FFFC9E}0.069 & \cellcolor[HTML]{FFFC9E}94.6     & \cellcolor[HTML]{FFFC9E}98.4     & \cellcolor[HTML]{FFFC9E}99.3     \\ \hline
\end{tabular}
\caption{Ablation study of different pipeline configurations. We change the backbone and compare various fusion strategies.}
\label{tab:ablation}
\end{table*}

\subsection{Ablation Study}
As an ablation study, we train our network with different configurations to show their impact on the final results. Detailed qualitative results on the individual setups can be found in the supplementary material. First, we trained with two different architectures that provide different bottleneck resolution (MonoDepth2\cite{monodepth2}: 1/32 resolution bottleneck, ToFNet\cite{su}: 1/4 resolution bottleneck). We compare with and without fusion to see the impact of implicit alignment with regard to the bottleneck resolution and the global context for phase unwrapping. Results in Tab.\ref{tab:ablation} indicate that without fusion, using 1/32 resolution bottleneck (ToFNet, No, I-ToF) already shows an improvement over 1/4 resolution bottleneck (MD2, No, I-ToF). The improvement is mostly on the phase unwrapping parts which implies the importance of the global context for this task in challenging scenes. Unwrapping related artefacts are reduced significantly, while the artefacts on the area where I-ToF is not reliable still remain. These can be removed with the additional RGB input. With fusion in the bottleneck, both architectures show an improvement, however, unwrapping related artefacts still exist after fusion with a 1/4 resolution bottleneck. Most of these artefacts are removed after fusion with the 1/32 resolution bottleneck. We argue that a smaller perceptive field combined with larger displacement in pixels at a higher resolution makes it harder to fuse and combine information in the former case.

For bottleneck fusion, we compare an attention based fusion which does not rely on an implicit alignment at low resolution. We first add a few layers on the bottleneck of RGB and I-ToF features and flatten them to obtain query, key, and value similar to \cite{attention1}. Then we aggregate the values with a weight obtained from the dot product between cross modal query and key. Result in Tab.\ref{tab:ablation} show that attention based fusion (MD2, Bottleneck(attention)) has some improvement over no fusion (MD2, No) while it does not perform as good as our implicit alignment strategy (MD2, Bottleneck(implicit align)). We believe that a huge domain difference between two features cannot be resolved with a few layers and a simple attention mechanism. In the supplementary material, we show details of the obtained attention map in three scenarios: I-ToF vs RGB, I-ToF vs I-ToF (temporal), RGB vs RGB (temporal) to support this argument.

At last, we train our network with and without explicit alignment to show the advantage of sharing higher resolution but aligned feature maps. As shown in Tab.\ref{tab:ablation}, quantitatively fusing higher resolution features (MD2, All resolutions) does not show a significant difference in comparison to bottleneck fusion (MD2, Bottleneck (implicit align)), however, it improves details such as texture copy artefacts which exist in the RGB depth branch, which we show this improvements later in the supplementary material.

\section{Conclusion}

We present a network architecture capable of dealing with realistic yet challenging scenarios for single frequency I-ToF where signals are acquired under strong ambient light with distances beyond the manufactures maximum depth range. Our pipeline uses both I-ToF and RGB information in a specially designed network which can incrementally resolve misalignments between the images while fusing their information from a low resolution. Our results show more than 40\% RMSE improvement over the state of the art raw I-ToF depth pipeline and around 10\% improvement over RGB guided depth completion. We strongly believe that this work can pave the way towards wider use of I-ToF in unconstrained environments and our fusion strategy of implicit alignment with follow-up explicit refinement can be a general contribution for spatially separated sensors fusion for depth estimation beyond ToF and RGB. Further work could exploit the inter-modal disparity of RGB and raw I-ToF (or depth respectively) to constrain the search space for the attention based fusion by epipolar geometry.

{\small
\bibliographystyle{ieee_fullname}
\bibliography{references}
}

\end{document}